\documentclass{article}
\pdfoutput=1
\date{}

\usepackage{enumitem}
\usepackage{spconf,amsmath,graphicx}
\usepackage{comment}
\usepackage{url}
\usepackage{hyperref} 
\usepackage[ruled,vlined]{algorithm2e}
\usepackage{multirow}
\usepackage{multicol}
\usepackage{float}

\begin{document}

\onecolumn 

\begin{description}[leftmargin=2cm,style=multiline]

\item[\textbf{Citation}]{Y. Sun, M. Prabhushankar, and G. AlRegib, "Implicit Saliency in Deep Neural Networks," in IEEE International Conference on Image Processing (ICIP), Abu Dhabi, United Arab Emirates, Oct. 2020.}

\item[\textbf{Review}]{Date of publication: 25 Oct 2020}

\item[\textbf{Codes}]{\url{https://github.com/olivesgatech/Implicit-Saliency}}

\item[\textbf{Copyright}]{\textcopyright 2020 IEEE. Personal use of this material is permitted. Permission from IEEE must be obtained for all other uses, in any current or future media, including reprinting/republishing this material for advertising or promotional purposes,
creating new collective works, for resale or redistribution to servers or lists, or reuse of any copyrighted component
of this work in other works. }

\item[\textbf{Contact}] {\href{mailto:ysun465@gatech.edu}{ysun465@gatech.edu} OR \href{mailto:mohit.p@gatech.edu}{mohit.p@gatech.edu}  OR \href{mailto:alregib@gatech.edu}{alregib@gatech.edu}\\ \url{http://ghassanalregib.com/} \\ }
\end{description}

\thispagestyle{empty}
\newpage
\clearpage
\setcounter{page}{1}

\twocolumn


\title{Implicit Saliency in Deep Neural Networks}
%
\name{Yutong Sun, Mohit Prabhushankar, and Ghassan AlRegib}
\address{OLIVES at the Center for Signal and Information Processing\\ School of Electrical and Computer Engineering\\ Georgia Institute of Technology, Atlanta, GA, 30332-0250\\ \{ysun465, mohit.p, alregib\}@gatech.edu}
%
%
%
\ninept
\maketitle
\begin{abstract}
In this paper, we show that existing recognition and localization deep architectures, that have not been exposed to eye tracking data or any saliency datasets, are capable of predicting the human visual saliency. We term this as \emph{implicit saliency} in deep neural networks. We calculate this \emph{implicit saliency} using expectancy-mismatch hypothesis in an \emph{unsupervised fashion}. Our experiments show that extracting saliency in this fashion provides comparable performance when measured against the state-of-art \emph{supervised} algorithms. Additionally, the robustness outperforms those algorithms when we add large noise to the input images. Also, we show that semantic features contribute more than low-level features for human visual saliency detection.
Based on these properties and performances, our proposed method greatly lowers the threshold for saliency detection in terms of required data and bridges the gap between human visual saliency and model saliency.
\end{abstract}
\begin{keywords}
Saliency, Implicit Saliency, Expectation Mismatch, Recognition, Deep Learning
\end{keywords}
\vspace{-3mm}
\section{Introduction}\vspace{-2mm}
\label{sec:intro}
\begin{figure*}[!htb]
\minipage{1\textwidth}
  \centering
  \includegraphics[width=0.7\linewidth,height=0.4\linewidth]{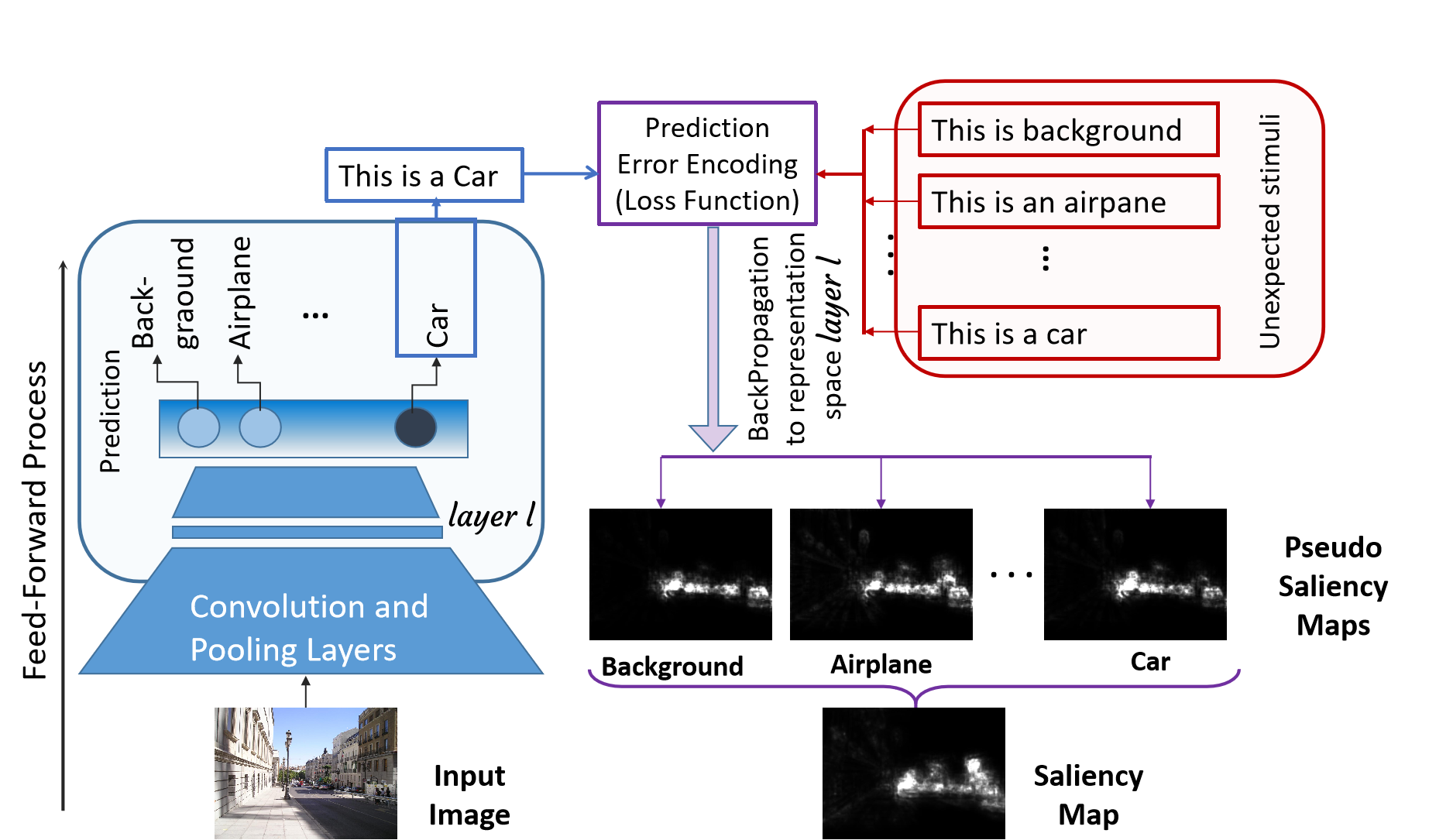}
\endminipage
\caption{Implicit Saliency Generation Process: We give an input image to a pretrained network and get an output vector based on prior knowledge of the network. We provide an unexpected stimuli which is a vector conflict to the output vector. We use a loss function to encode the error based on the output vector and the conflict vector. Backpropagating this error to a semantic convolutional layer results in pseudo saliency maps. We combine the resultant pseudo saliency maps using statistical methods to obtain the final saliency map.}\label{fig:Implicit
Saliency}
\end{figure*}

Saliency is defined as those regions in a visual scene that are `most noticeable' and attract significant attention~\cite{uddin2016salience}. 
Human visual saliency detection has been deployed in an extensive set of image processing applications including but not limited to data compression, image segmentation, recognition, image quality assessment (IQA) and object recognition~\cite{alshawi2018uncertainty}. Broadly, saliency detection algorithms can be classified into two categories. The first is bottom-up approaches where saliency detection techniques extract features from  data and compute saliency based on extracted features~\cite{DBLP:journals/corr/KummererWB16, cornia2018sam, cornia2016deep}. The second is top-down approaches where the algorithms have a prior target for which features are to be calculated ~\cite{murabito2018top}. Both these approaches derive from the expectancy mismatch hypothesis~\cite{BECKER2011290}.     

The expectancy-mismatch hypothesis for a sensory system is based on receiving information which is in conflict with the system's prior expectation. The authors in~\cite{BECKER2011290} show that a message which is unexpected, captures human attention and is hence salient. Extensive work in the field of cognitive sciences has been conducted to study the impact of expectancy-mismatch in human attention and visual saliency~\cite{SUMMERFIELD2009403,krebs2012stimulus,horstmann2016perceptual,doi:10.1111/1467-9280.00488,BECKER2011290}. Based on these works, human attention mechanism suppresses expected messages and focuses on the unexpected ones. During this process, human visual system checks whether the input scenario matches the observers' expectation and past experience. When they are conflicting, error neurons in human brain encode the prediction error and pass the error message back to the representational neurons. Existing work applies this concept of expectancy-mismatch to saliency detection. The authors in~\cite{horstmann2016perceptual,doi:10.1111/1467-9280.00488} show how unexpected colors impact human eye fixations. \cite{BECKER2011290} indicates that a motion singleton captures attention.

Previous works that define expectations and calculate mismatches are based on low-level representations like colors and edges. However, the advent of deep learning has shown the importance of semantic information that combines low-level features for complicated tasks like recognition. Neural networks have shown an aptitude for learning higher-order semantic representations. In~\cite{sun2018semantic}, the authors claim that it is crucial to consider semantic representations in saliency detection. In this paper, we propose to create expectancy based on high level semantic features and calculate mismatch from input information to obtain saliency. To set expectancy, we use neural networks.
To calculate mismatch, we provide conflicting information to the network along with the input image to search for those regions in the input image that are affected by the conflict. In this work, conflicting information refers to labels that conflict with predicted classes. For instance, consider Fig.~\ref{fig:Implicit Saliency}. The network has learned the low-level features like edges and colors and their combinatorial high-level semantics to recognize a car. However, by providing a conflicting label such as `airplane', we force the network to reexamine its decision process. The network reconciles its expectation of finding a car and the conflicting label that it is an airplane by encoding the error within the gradients. These gradients are backpropagated throughout the network to resolve the conflict. The change brought about by the gradients is indicative of regions within the image that are used for expecting the output. We postulate that these regions are thereby salient.

In this work, we use commonly used recognition and localization pre-trained networks to set expectancy. These networks have not been exposed to either saliency datasets or eye-tracking data. Hence, the proposed method is completely unsupervised. We extract saliency that is implicitly embedded within any given network. Hence, the proposed approach is termed implicit saliency in neural networks. The contributions in this paper are three-fold:
\begin{enumerate}
\setlength{\itemsep}{0ex}
\setlength{\parsep}{0ex}
\setlength{\parskip}{0.5ex}
    \item[1)] we extract implicit saliency from pre-trained networks that have not seen eye-tracking data in an \emph{unsupervised} fashion.
    \item[2)] we show that the proposed implicit saliency is robust to noise.
    \item[3)] we show that semantic features combined with unexpected stimuli have a higher correlation with human visual saliency than low-level features or semantic features without unexpected stimuli.
\end{enumerate}

We introduce the background for the pre-trained deep neural networks in Section~ \ref{sec:Background}. In Section~\ref{sec:Method}, we detail the proposed method to extract implicit saliency. In Section~\ref{sec:Application}, we compare the performance of proposed method against state-of-art supervised methods and model saliency methods. We conclude in Sec.~\ref{sec:Conclusion}.

\vspace{-4mm}
\section{Background}
\label{sec:Background}
\vspace{-1mm}
Visual recognition is a common activity on which humans heavily rely to interact within their environments both accurately and rapidly. It has been shown that the process of recognition and categorization inside humans takes place within $300\text{ms}$~\cite{DICARLO2007333,thorpe1996speed}. Furthermore, this process is performed in the cortex area of the brain, which is controlled by human attention mechanism. Hence recognition highly relates to human attention selection~\cite{doi:10.1152/jn.00777.2002}. Therefore, in this paper recognition networks are utilized as the backbone of the proposed implicit saliency as shown in Fig.~\ref{fig:Implicit Saliency}. Specifically, we utilize ResNet-$18$, $34$, $50$, $101$~\cite{he2016deep}, and VGG$16$~\cite{Simonyan15} that are pre-trained on ImageNet~\cite{deng2009imagenet} as well as Faster R-CNN~\cite{renNIPS15fasterrcnn} that is pre-trained on PASCAL VOC 2007+2012~\cite{Everingham10}. We implement this project in PyTorch.

We denote an $L$ layered network as $f()$, its weights as $W$, and bias as $b$.  
During the training process, $W$ and $b$ are updated until the model is parameterized by these these weights and bias. A pre-trained network provides the expectation $y_{pred}$ based on the prior knowledge obtained during the feed-forward process. We call such features  as \emph{feed-forward features}. We also denote the provided conflicting information mentioned in Sec.~\ref{sec:intro}  as $y_{unexp}$ and the gradients corresponding to the encoding error as \emph{conflicting features}. 

\vspace{-4mm}
\section{Implicit Saliency Generation}\vspace{-1mm}
\label{sec:Method}
\begin{figure}[!h]\vspace{-3mm}
\centering
  \includegraphics[width=\linewidth]{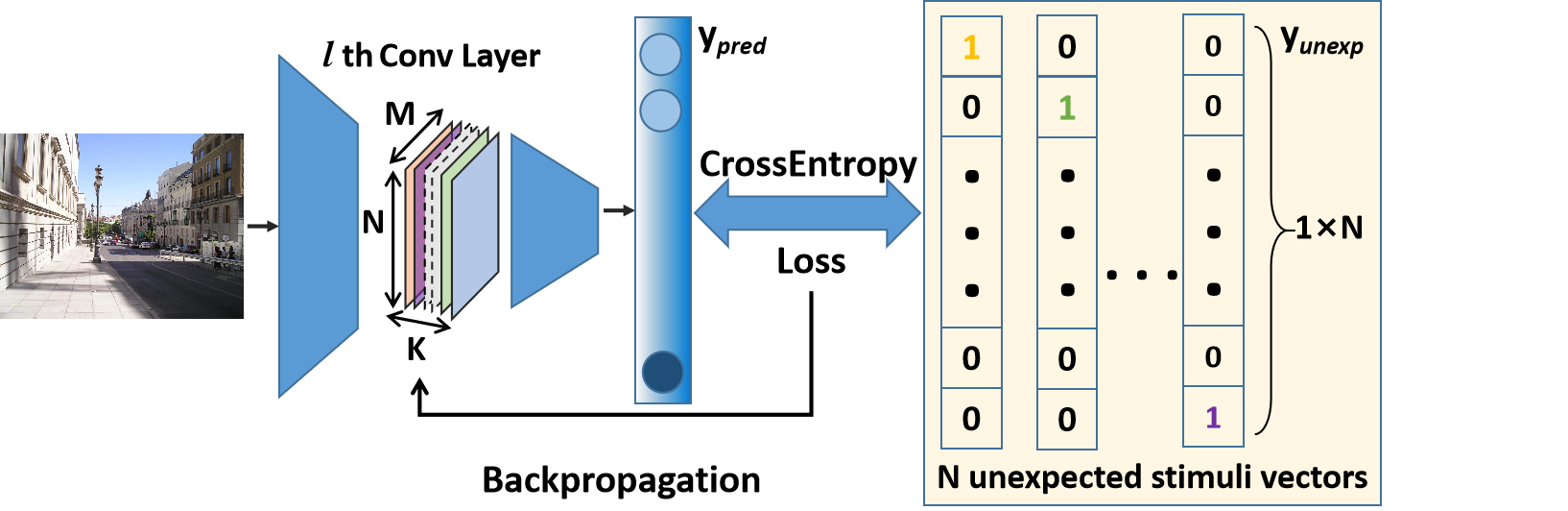}
\vspace{-5mm}
\caption{Conflicting feature generation}
\label{fig:Conflictingfeaturegeneration}\vspace{1mm}
\end{figure}
\begin{figure*}[!h]
\centering
\minipage{1\textwidth}%
  \includegraphics[width=\linewidth]{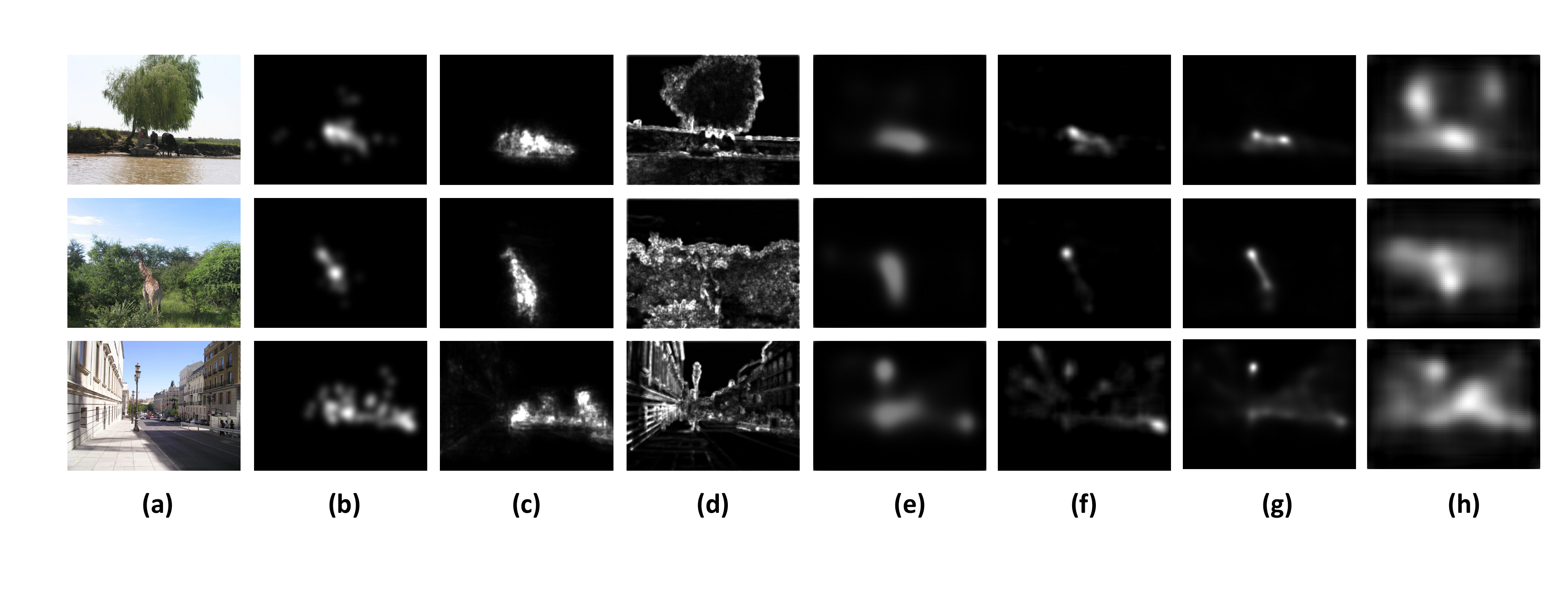}
\endminipage
\vspace{-10mm}
\caption{Saliency map visualization. (a) Input image (b) Groudtruth (c) Proposed Method (d) Feed-forward feature (e) SalGan~\cite{pan2017salgan} (f) ML-Net~\cite{cornia2016deep} (g) DeepGazeII~\cite{kummerer2016deepgaze} (h) ShallowDeep~\cite{Pan_2016_CVPR}}
\label{fig:Saliency_Visualizaiton}
\vspace{-3mm}
\end{figure*}

\begin{figure*}[!h]\vspace{-3mm}
\centering
\minipage{1\textwidth}%
  \includegraphics[width=\linewidth]{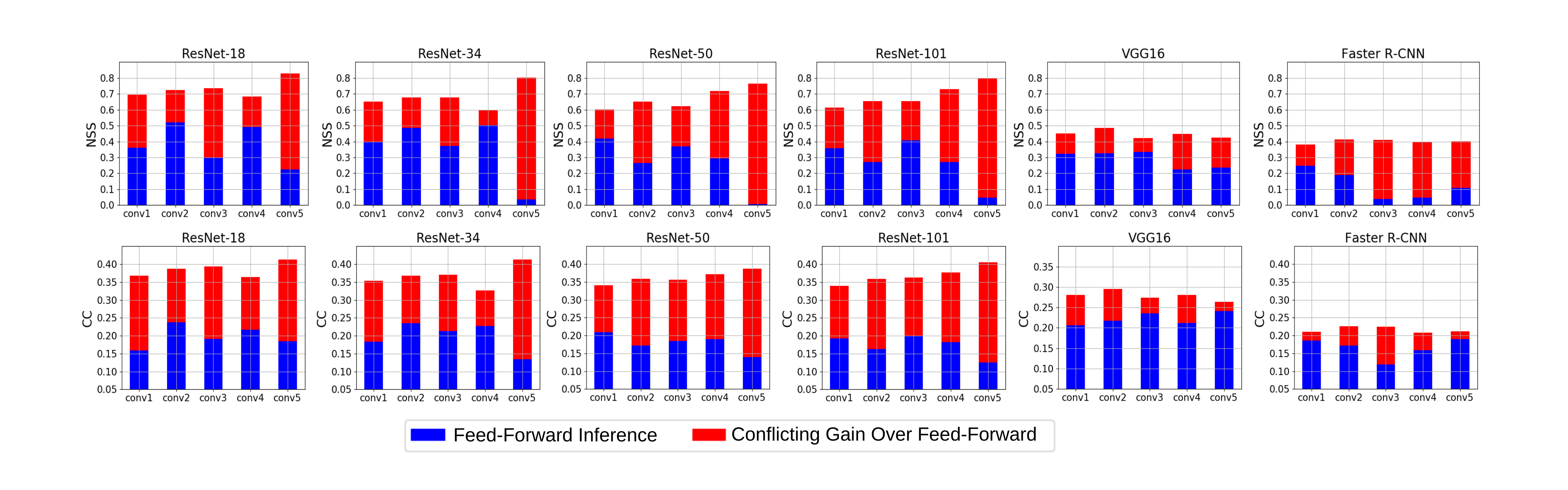}
\endminipage
\vspace{-5mm}
\caption{Visualization of NSS and CC gains (in red) of using the proposed conflicting features over feed-forward features on MIT1003}\label{fig:Saliency_Validation}\vspace{-3mm}

\end{figure*}


Let the pre-trained network $f()$ have $N$ classes. Therefore, the prediction $y_{pred}$ and the unexpected stimulus $y_{unexp}$ are $1 \times N$ vectors. Each class $i$ has a corresponding one-hot unexpected stimulus vector $y_{unexp}^{i}$.  As shown in Fig.~\ref{fig:Conflictingfeaturegeneration}, having $y_{pred}$ and $y_{unexp}^{i}$ in hand, we encode the unexpected information by a convex loss function $J()$. The encoded unexpected information is denoted as $J(W,x,i)$, where $W$ is the weight, $x$ is the input, and $i$ is the assigned class in the unexpected stimulus. Note that this formulation pf $J()$ is similar to the one described in~\cite{kwon2019distorted}. In this paper, we use CrossEntropy function for $J()$. The generation process of proposed implicit saliency map is shown in Alg.~\ref{algorithm:ImplicitSaliency}. 
$R$ represents the gradients to resolve the conflict on a specific semantic layer. 
Since there are $N$ classes in this pre-trained network, we get $N$ pseudo-saliency maps. Each pseudo-saiency map shows the salient region corresponding to the class $i$ in the given unexpected stimulus. Notice that we also take the network's decision class as an unexpected stimulus because the decision comes from the highest score in the output, but the network is still not $100\%$ sure with its decision thereby providing a non-zero loss $J()$. 

\begin{algorithm}[!h]
\caption{Implicit Saliency Map Generation}
\begin{enumerate}
\setlength{\itemsep}{0ex}
\setlength{\parsep}{0ex}
\setlength{\parskip}{0.5ex}
\item[\textbf{Step 1:}] Generate pseudo saliency maps on a specific convolution layer as:\\
$R_{i,m,n}^k=abs(\nabla_{f_{k,l}}(J(W_{m \times n},x,i)))$\\
$f$: feature maps corresponding to the convolution filter\\
$k \in [1, K]$: index for each convolution filter\\
$l\in [1,L]$: specific convolution layer\\ $m \in [1,M], n \in [1,N]$: spatial indices
\item[\textbf{Step 2:}] Average pseudo saliency maps over class dimension:
$\mu_{m,n}^k= \frac{1}{N}\sum_{i=1}^{N} abs(\nabla_{f_{k,l}}(J(W_{m \times n},x,i)))$
\item[\textbf{Step 3:}] Generate the variance mask over class dimension:
$V_{m,n}^k=\frac{1}{N}\sum_{i=1}^{N}(R_{i,m,n}^k-\mu_{m,n}^k)^2$
\item[\textbf{Step 4:}] Generate implicit saliency map:\\
$S_{C}=\frac{1}{K}\sum_{k=1}^{K}(1-\hat{V}_{m,n}^k)\circ \mu_{m,n}^k$\\
$\circ$: element-wise multiplication\\$\hat{V}$:normalized variance map. 
\end{enumerate}
\vspace{-1mm}
\label{algorithm:ImplicitSaliency}
\end{algorithm}


 As shown in Fig.~\ref{fig:Implicit Saliency}, for different classes, pseudo-saliency maps focus on different salient regions. The final implicit saliency map is the combination of all these salient regions. Specifically, we use mean to combine all the pseudo saliency maps. To decrease the uncertainty in the overall saliency map, we negate the variance of all the pseudo saliency maps, which in shown in Step 4 of Alg.~ \ref{algorithm:ImplicitSaliency}. Qualitatively, we can see that the right part of the implicit saliency map is primarily influenced by the pseudo-saliency maps from airplane-like classes while the left part is derived from the car-like classes. Note that each pseudo-saliency map is a matrix with the same dimensionalities as the convolution filters in $l$th layer. In Fig.~\ref{fig:Implicit Saliency}, we sum it up over depth dimension and rescale it to $[0,255]$ for visualization.

\vspace{-4mm}
\section{Experiments}
\vspace{-1mm}
\label{sec:Application}
In Section~\ref{sec:intro}, we motivate implicit saliency using expectancy-mismatch. We generate saliency maps by highlighting regions where the network is re-examining its decision process because of provided conflict. In this section, we sequentially validate these arguments. First, we demonstrate that expectancy-mismatch in neural networks has a higher correlation with saliency as compared to feed-forward expectancy features. As such we compare the proposed implicit saliency against feed-forward feature maps. Next, we show that the regions that are in conflict with the decision process are more salient than the regions used to make the decision. Finally, we compare the performance and robustness of proposed unsupervised implicit saliency against state-of-the-art supervised saliency detection methods.

\vspace{-1mm}
\subsection{Implicit saliency of pre-trained networks}
\vspace{-1mm}
The expectancy of an input image is encoded in the feed-forward activation maps. In this experiment, we compare the saliency maps obtained by feed-forward expectancy features and proposed expectancy-mismatch features. In the feed-forward method, we use the same statistical process as Alg.~\ref{algorithm:ImplicitSaliency} and denote the saliency map as $S_F$. Note that instead of using gradients, we use activations to obtain $S_F$. The saliency map based on conflicting features is $S_C$ shown in 
Sec.~\ref{sec:Method}. We validate the saliency detection capabilities of feed-forward and conflicting features on MIT1003~\cite{mit-saliency-benchmark} dataset. MIT1003 consists of $1003$ images and the corresponding eye tracking data from human subjects. We qualitatively evaluate the performance of saliency detection in Fig.~\ref{fig:Saliency_Visualizaiton}. The feed-forward features in Fig.~\ref{fig:Saliency_Visualizaiton}d. focus on edges and textures without specific localization. The proposed implicit saliency generates localized saliency maps which are highly correlated with the ground truth. These visualizations show that the visual saliency is more effectively captured through the expectancy-mismatch process than expectancy process.

\begin{table*}[!tb]
\centering
\caption{Human visual saliency vs Model Saliency}
\small
\begin{tabular}{|c|c|c|c|c|c|c|c|c|c|}
\hline
\multirow{2}{*}{} &
\multicolumn{4}{c|}{NSS} &
\multicolumn{4}{c|}{CC} \\
\hline
Networks&ResNet-18 & ResNet-34 & ResNet-50 &ResNet-101& ResNet-18 & ResNet-34 & ResNet-50 &ResNet-101 \\
\hline
GradCam & $0.7657$
 & $0.7545
$ & $0.7203
$ &$0.7335
$& $0.3496
$ & $0.3396
$ & $0.3190
$ &$0.3210
$ \\
\hline
GBP & $0.3862
$ & $0.4191
$ & $0.3898
$ &$0.3415
$& $0.2474
$ & $0.2453
$ & $0.2443
$ &$0.2233
$ \\
\hline
\textbf{ImplicitSaliency} & $\mathbf{0.8274
}$ & $\mathbf{0.8018
}$ & $\mathbf{0.7659
}$ &$\mathbf{0.7981
}$ & $\mathbf{0.4132
}$ & $\mathbf{0.4112
}$ & $\mathbf{0.3868
}$ &$\mathbf{0.4051
}$ \\
\hline
\end{tabular}
\label{table:Model Saliency}
\end{table*}

\begin{table*}[!tp]
\small
\centering
\caption{Robustness Analysis of Implicit Saliency}
\begin{tabular}{|c|c|c|c|c|c|c|c|c|c|c|c|}
\hline
\multirow{2}{*}{} &
\multicolumn{5}{c|}{NSS} &
\multicolumn{5}{c|}{CC} \\
\hline
 Gaussian &Sal&Deep&ML&Shallow&\textbf{Implicit} &Sal&Deep&ML&Shallow&\textbf{Implicit}\\ [0ex] 
 Blur   &Gan&GazeII&Net&Deep&\textbf{Saliency} &Gan&GazeII&Net&Deep&\textbf{Saliency}\\ [0ex] 
\hline
$r=0$ & $0.8977$ & $0.6214$ & $0.5431$ &$\mathbf{0.9306}$&$0.7981$ & $\mathbf{0.6280}$ & $0.5927$ &$0.4481$&$0.5120$ &$0.4051$\\
\hline
$r=50$ & $\downarrow0.2239$ & $\downarrow0.3436$ &$\downarrow0.2484$ &$\downarrow0.2025$& $\mathbf{\downarrow0.1793}$&$\downarrow0.2731$ &$\downarrow0.3954$ &$\downarrow0.2940$&$\downarrow0.1840$&$\mathbf{\downarrow0.1432}$ \\
\hline

\end{tabular}
\label{table:NSS and CC result}
\end{table*}

We also quantitatively evaluate both methods using Normalized Scanpath Saliency (NSS) and Correlation Coefficient (CC) and report results in Fig.~\ref{fig:Saliency_Validation}. Based on~\cite{bylinskii2018different}, NSS computes the average normalized saliency at fixation locations. High positive NSS indicates high correspondence, while negative NSS indicates anti-corresponds. CC measures the correlation between saliency maps and ground truth. Higher CC indicates better performance. Since different layers extract different semantic information, we show results from $5$ convolution layers for each model. Based on Fig.~\ref{fig:Saliency_Validation}, the proposed method outperforms the feed-forward method by $0.32$ on average for NSS, and $0.14$ for CC. Also, the proposed method achieves more robust performance over different layers compared to the feed-forward feature. The maximum performance drop of the feed-forward feature method is $98\%$ and $42\%$ for NSS and CC across different layers. The maximum drop in conflicting feature is only $25\%$ and $21\%$ for NSS and CC, which shows our proposed method achieves stable saliency detection results across layers. These results validate the usage of expectancy-mismatch in semantic layers as compared to feed-forward features in the same layers.

\vspace{-4mm}
\subsection{Implicit Saliency vs Model Saliency}
\vspace{-1mm}
In this experiment, we show that regions where the network re-examines its decision are more salient compared to regions that are used to make decisions. Regions that are used to make decisions in a recognition network are obtained using model saliency. In this paper, we consider two model saliency methods - Grad-CAM~\cite{selvaraju2017grad} and Guided Backpropagation (GBP)~\cite{springenberg2014striving}. These methods are compared against implicit saliency among ResNet-18,34,50,101 architectures. Note that both these model saliency methods are unsupervised methods. Table~\ref{table:Model Saliency} shows the proposed method outperforms both GradCam and GBP in both NSS and CC metrics. GBP shows the lowest results since it is designed to find only low-level features like edges. Meanwhile, Grad-CAM performs relatively well since it focuses on semantic features in the given image. However, Grad-CAM only uses the model's decision as guidance to find the salient regions. Our proposed method utilizes the unexpected stimuli to extract high-level semantic features based on expectancy-mismatch hypothesis. From this experiment we can see that while low-level features from GBP are important for a network to make its decision, they have relatively low correlation with human attention. Semantic features that are important for the network to make decision are correlated to human attention. This correlation can be increased by using proposed expectancy-mismatch.

\vspace{-3mm}
\subsection{Robustness Analysis of Implicit Saliency}
\vspace{-1mm}
In this experiment, we compare our proposed method with 4 state-of-art methods: SalGan~\cite{pan2017salgan}, DeepGazeII~\cite{kummerer2016deepgaze}, ML-Net~\cite{cornia2016deep} and Shallow and Deep Networks~\cite{Pan_2016_CVPR}. All these models are trained on SALICON~\cite{jiang2015salicon}. SALICON is an eye-tracking dataset that offers large number of saliency annotations on commonly used datasets like MS-COCO~\cite{lin2014microsoft}. We visualize the qualitative resulst of all these methods in Fig.~\ref{fig:Saliency_Visualizaiton} (e), (f), (g), (h). The results from~\cite{Pan_2016_CVPR} visualized in Fig.~\ref{fig:Saliency_Visualizaiton}(h) cover a more comprehensive area, while Fig.~\ref{fig:Saliency_Visualizaiton}(e), (f) and (g) provide a higher precision. The proposed implicit saliency Fig.~\ref{fig:Saliency_Visualizaiton}(c) is both fine-grained and precise. 

We quantitatively compare these methods against the proposed implicit saliency in Table~\ref{table:NSS and CC result}. To ascertain robustness of all the methods, we add Gaussian blur with $radius=50$ to input images from MIT1003. When there is no noise in the input images, our proposed method is the third best for NSS and the fourth best for CC among these $5$ methods. With noise added, our proposed method's performance drops the least. For NSS, our performance drop is $0.075$ lower than the state-of-art algorithms on average, and $0.164$ lower than the largest drop. For CC, our performance drop is $0.143$ lower on average, and $0.252$ lower than the largest drop. Note that the $4$ comparison methods are supervised networks while the proposed method is completely unsupervised. The comparison methods are well-trained on complex scenarios similar to MIT1003. However, those scenarios are not common in PASCAL VOC 2007+2012 and ImageNet that are used to train networks in our proposed method. Also, the proposed method's pre-trained networks are trained for tasks other than saliency detection. Despite these handicaps, the proposed implicit saliency based on expectancy-mismatch in semantic information, provides a comparable performance to supervised networks based on both qualitative and quantitative results. 

\vspace{-4mm}
\section{Conclusion}
\vspace{-1mm}
\label{sec:Conclusion}
In this paper we propose implicit saliency that is extracted from a pre-trained network based on expectancy-mismatch hypothesis. This network can be any of classification, detection, or recognition networks. Based on three experiments, we show that our proposed method has higher correlation with human visual saliency than only using feed-forward features. Our method is also stable across layers. The proposed implicit saliency is achieves a comparable performance and is more robust when measured against state-of-the-art \emph{supervised} saliency detection methods. Additionally, by comparing with two model saliency methods, we show that our unexpected based semantic saliency features have higher correlation with human visual saliency than low-level features and semantic features without unexpected stimuli. Our method is completely \emph{unsupervised}. This greatly lowers the threshold for saliency detection in terms of data collection. Also, existence of implicit saliency in neural networks can bridge the gap between recognition and neuroscience communities. Human visual saliency can be shown to be embedded into all neural networks thereby increasing the understanding of both saliency and neural networks.
\vspace{-4mm}

\ninept
\bibliographystyle{IEEEbib}
\bibliography{strings,refs}

\end{document}